\documentclass{article}

\usepackage{arxiv}

\usepackage[utf8]{inputenc} 
\usepackage[T1]{fontenc}    
\usepackage{hyperref}       
\usepackage{url}            
\usepackage{booktabs}       
\usepackage{amsfonts}       
\usepackage{nicefrac}       
\usepackage{microtype}      
\usepackage{lipsum}
\usepackage{graphicx}
\graphicspath{ {./images/} }

\title{Fake News Detection in Spanish Using Deep Learning Techniques}

\author{
  Kevin Mart\'inez-Gallego \\
  Intelligent Information Systems Lab\\
  Universidad de Antioquia\\
  Calle 67 No. 53 - 108, 050010, Medell\'in, Colombia. \\
  \texttt{kevin.martinez@udea.edu.co} \\
   \And
 Andr\'es M. \'Alvarez-Ortiz \\
  Intelligent Information Systems Lab\\
  Universidad de Antioquia\\
  Calle 67 No. 53 - 108, 050010, Medell\'in, Colombia. \\
  \texttt{amauricio.alvarez@udea.edu.co} \\
  \And
  Juli\'an D. Arias-Londo\~no \\
  Intelligent Information Systems Lab\\
  Dpt. of Systems Engineering and Computer Science\\
  Universidad de Antioquia\\
  Calle 67 No. 53 - 108, 050010, Medell\'in, Colombia. \\
  \texttt{julian.ariasl@udea.edu.co} \\
}

\begin{document}
\maketitle
\begin{abstract}

This paper addresses the problem of fake news detection in Spanish using Machine Learning techniques. It is fundamentally the same problem tackled for the English language; however, there is not a significant amount of publicly available and adequately labeled fake news in Spanish to effectively train a Machine Learning model, similarly to those proposed for the English language. Therefore, this work explores different training strategies and architectures to establish a baseline for further research in this area. Four datasets were used, two in English and two in Spanish, and four experimental schemes were tested, including a baseline with classical Machine Learning models, trained and validated using a small dataset in Spanish. The remaining schemes include state-of-the-art Deep Learning models trained (or fine-tuned) and validated in English, trained and validated in Spanish, and fitted in English and validated with automatic translated Spanish sentences. The Deep Learning architectures were built on top of different pre-trained Word Embedding representations, including GloVe, ELMo, BERT, and BETO (a BERT version trained on a large corpus in Spanish). According to the results, the best strategy was a combination of a pre-trained BETO model and a Recurrent Neural Network based on LSTM layers, yielding an accuracy of up to 80\%;
nonetheless,  a baseline model using a Random Forest estimator obtained similar outcomes. Additionally, the translation strategy did not yield acceptable results because of the propagation error; there was also observed a significant difference in models performance when trained in English or Spanish, mainly attributable to the number of samples available for each language.\\
\end{abstract}


\section{Introduction}

In social networks, the proliferation of fake news is a strategy used to manipulate public opinion. For example, it is well-known the case of Cambridge Analytica, where the ``private data of millions of people was used to psychologically manipulate voters in the 2016 US elections, where Donald Trump was elected president. The company not only sent tailored advertising but developed fake news that it then replicated across social networks, blogs and media" \cite{bbc_news_2018}. One of the strategies that have begun to be explored to prevent the proliferation of fake news, is the use of Artificial Intelligence (AI) techniques and, more precisely, Deep Learning (DL) for their detection and subsequent removal. Most of the work that has been done in this field uses datasets of news written in English as a source of information, which are composed of properly labeled sentences publicly available.  Although fake news is a common problem across different languages, including Spanish, there is not a significant amount of properly labeled fake news in Spanish to effectively train a DL model for fake news detection, similar to those proposed for the English language. Indeed, to the best of our knowledge, recently in 2019 the first corpus of fake news in Spanish exclusively adapted for such a task was presented in \cite{posadas2019detection}; nevertheless, this corpus consists of 971 labeled news, which is an insufficient amount of samples to train a solid DL model from scratch. Therefore, the main objective of this work is to design a Machine Learning (ML) strategy for the detection of fake news in Spanish, based on Transfer Learning techniques and/or machine translation tools, which allow the use of previously trained models, both in English and Spanish.\\

This paper is organized as follows: section \ref{sec:related_work} presents some antecedents on the use of ML and DL models for fake news detection in different languages, doing emphasis on English and Spanish; section \ref{sec:methods} presents the pre-processing strategies applied to the texts, and also the models and embeddings we employed; in section \ref{sec:experiments_and_results} we present the datasets utilized and the methodology for evaluating the different models, as well as the settings for the experiments carried out and the outcomes we obtained. Finally, we discuss the results and present the conclusions of this paper in section \ref{sec:discussion_and_conclusions}.

\section{Related Work}
\label{sec:related_work}

Automatic fake news detection (FND) is a task that has attracted extensive attention from AI researchers in recent years; this has been evidenced by the large number of publications in which the problem has been addressed by applying different strategies. Shu \textit{et al.} \cite{shu2017fake} report a summary of research works on FND in social networks, analyzing the aspects involved from psychology, social theories, and algorithmic points of view. As in many ML applications, the proposed approaches addressing FND are composed of two stages: feature extraction and model building; the first refers to the numerical representation of news content and related information; the second proposes the development of a machine learning model to distinguish between fake news and legitimate news. For example, Wu and Liu in \cite{wu2018tracing} assume that fake news is typically manipulated to resemble real news; thus, they propose a classifier based on propagation paths in social networks using Long Short-Term Memory Recurrent Neural Networks (LSTM-RNN) and Embeddings. Although the FND task has traditionally been stated as a bi-class classification problem, in \cite{wang2017liar} the author presents a dataset in English (\emph{The Liar Dataset}), which is composed of 6 classes: pants-fire, false, barely true, half-true, mostly true, and true. In addition, this author evaluates four classification models following an approach where he considered both meta-data and text; hence, he presented Support Vector Machine (SVM) as the best classical ML model,  and Convolutional Neural Network (CNN) as the best DL model, which outperformed the other models with an accuracy of 27\% on the test set. Using this same dataset, Bracsoveanu and Andonie in \cite{bracsoveanu2019semantic}, propose to add a pre-processing stage based on the extraction of \textbf{semantic features} from the text. These authors also evaluate classical ML and DL models, finding the SVM model to be the best in terms of performance (28.4\%) for classical ML, and the CapNetLSTM model (64.4\%) for DL, which was used in combination with a pre-trained Embeddings model; these results were obtained on the dataset presented in \cite{wang2017liar}.  The authors conclude that employing \emph{Semantic Features} significantly improves accuracy in fake news detection; in particular, for DL models, the improvement in accuracy was up to 5-6\%. Furthermore, they also highlighted that "the accuracy of the various models greatly varies depending on the data sets and the number of classes involved", a phenomenon we also noticed across this state-of-the-art review.

Previous works tackled the FND task using datasets in English; however, this paper focuses on FND for Spanish. Faustini and Covoes propose in \cite{faustini2020fake} an approach using text features that can be generated independently of the news source platform and, as far as possible, independently of the news language under analysis. The authors report competitive results for news in languages belonging to the Germanic, Latin, and Slavic language groups. They used five datasets, and each one was processed with four different Natural Language Processing (NLP) techniques for text representation. Then, experiments were performed with different models obtaining the best result with Random Forest and SVM algorithms, combined with Bag-of-Words (BoW) as text representation technique; hence, they got a prediction rate of up to 95\% for news in the specified linguistic groups. Additionally, Posadas-Durán \textit{et al.} \cite{posadas2019detection} address the FND task for Spanish news, using different classical ML models: SVM, Random Forest, Logistic Regression, and Boosting; these models were combined with different strategies for text pre-processing that allow extracting useful semantic information for the detection task: BOW, Part of Speech tags (POS tags) and n-grams, as well as applying Stop Words to avoid prepositions and/or punctuation marks in the text. The experiments were carried out using a proprietary dataset \footnote{\url{https://github.com/jpposadas/FakeNewsCorpusSpanish}}  (released under CC-BY-4.0 license). The authors report results of up to 77.28\% accuracy for one of the combinations.

To the best of our knowledge, no works applying DL models in the Spanish FND task have been published so far.

\section{Methods}
\label{sec:methods}

\subsection{Preprocessing steps}

In order to obtain consistent results, a data standardization process known as \emph{Text Normalization} was performed, which, in addition to eliminating non-alphanumeric characters in the text, includes some of the most commonly used techniques in NLP:
\begin{itemize}
  \item \textbf{\emph{Stop Words:}} we removed words there is an agreement they do not contribute to the models learning process in the context of the problem addressed; for instance, articles and prepositions.
  \item \textbf{\emph{Stemming:}} this technique was used to reduce words to their root. 
  \item \textbf{\emph{Tokenization and Padding:}} as usual in text processing tasks, we performed tokenization and padding, when required, for words and sentences representation.
  
\end{itemize}

Subsequently, we decided to compare some of the most common techniques regarding text representation: BoW, which provides the number of occurrences of each word in the text corpus; \emph{term frequency-inverse document frequency} (tf-idf), which provides a weighted measure of the importance of each term within the text (according to its frequency of occurrence in sentences); and pre-trained \emph{Word Embeddings}, where words and the semantic relationships among them are represented as a vector. It is worth clarifying that, we call Word Embeddings to both pre-trained vectors such as word2vec or GloVe, and embeddings obtained from pre-trained models such as ELMo or BERT (presented in subsection \ref{sec:models}).

\subsection{Models}
\label{sec:models}

\begin{figure}[!htbp]
        \centering
        \includegraphics[width=0.55\columnwidth]{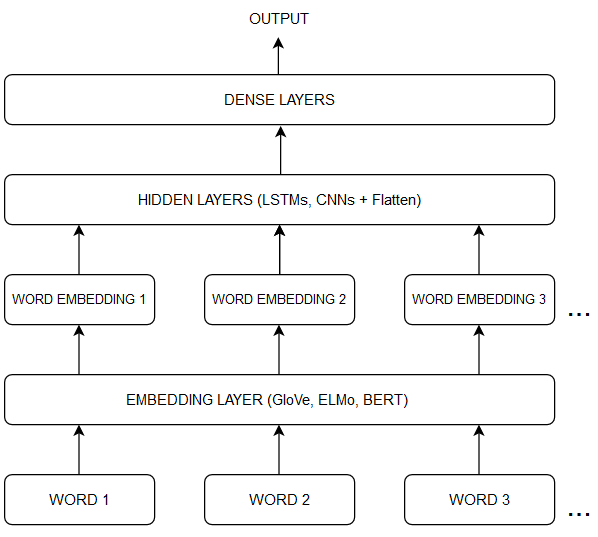}
        \caption{Schematic of DL classifiers in combination with Embedding Layer}
        \label{fig:DL_Architecture_Embedding}
\end{figure}

Classical ML models, and DL models based on artificial neural networks were used. We employed ML models intending to create a baseline for comparison purposes; hence, we selected the following: Support Vector Machine (SVM), Random Forest (RF), Gradient Boosting Tree (GBT), and Multi-Layer Perceptron (MLP).  For the case of DL classifiers, besides word embeddings, two types of layers were used: Long Short-Term Memory Recurrent Neural Network (LSTM-RNN) using a \emph{many-to-one} architecture, and Convolutional Neural Network (CNN). LSTM-RNN processes the input data as sequences of dependent observations, while CNNs can process n-grams through the application of convolutional filters. A schematic of the DL classifiers in combination with a embedding layer is illustrated in Figure \ref{fig:DL_Architecture_Embedding}; this figure shows the arrangement of the aforementioned layers, and the different word embeddings we used which are presented next.

In some experiments, we trained the embedding layer as part of the model training process. However, in most cases, we used transfer learning strategies through pre-trained word embeddings. This procedure makes much sense when either there are not many samples available for training or the computational resources are limited. Four word-embedding variants were evaluated:     

\begin{enumerate}

    \item \textbf{\emph{Global Vectors for Word Representation} (GloVe):} it is an unsupervised method that captures statistical information from a text corpus; in order to generate word representations, its training process is based on the spectral co-occurrence matrix decomposition \cite{pennington2014glove}. Hence, we made use of the 300-dimensional vectors pre-trained on the English Wikipedia corpus, available at \cite{pennington_2014}.
    
    \item \textbf{\emph{Embeddings from a Language Model} (ELMo):} in contrast to GloVe, which provides a fixed meaning for each word, the representations generated by ELMo are functions of the entire sequence, instead of a single word. ELMo encodes words by individual characters, such that it allows the same word to have different representative vectors under different contexts \cite{peters2018deep}. The pre-trained model was downloaded from the public repository Tensorflow Hub, which can be found at \cite{TensorFlow_Elmo}.
    
    \item  \textbf{\emph{Bidirectional Encoder Representations from Transformers} (BERT):} it is a language representation model based on another type of model called Transformer, where instead of strictly analyzing the temporal dependence of the input sequence, all possible combinations of the input sequence are evaluated through an attention layer \cite{devlin2018bert}. It has the advantage that its training process can be performed in parallel, since it does not depend on the temporal condition. For this research, the so-called \emph{BERT Base} was used, which is a model with a total of 110 million pre-trained parameters. The pre-trained model was downloaded from the public repository Tensorflow Hub, available at \cite{TensorFlow_Bert}.
    
    \item \textbf{BETO:} this model corresponds to a BERT version but instead of English, this models is trained on a large corpus of text in Spanish \cite{CaneteCFP2020}. The size of the model is similar to a BERT Base, with approximately 110 million parameters.
        
 \end{enumerate}

\section{Experiments and Results}
\label{sec:experiments_and_results}

\subsection{Datasets}

\begin{figure}[!htbp]
    \centering
    \includegraphics[width=0.55\columnwidth]{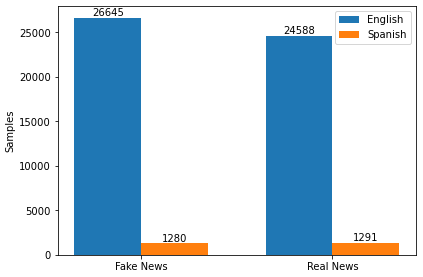}
    \caption{Sample distribution for the resulting datasets in English and Spanish}
    \label{fig:db_dist}
\end{figure}

Four free-to-use datasets were chosen to use in this study. Two of them consist of news in English labeled as fake or real: \emph{Fake and real news dataset} \cite{bisaillon_2020}, and \emph{News Data Set - Fake OR Real} \cite{ukani_2020}; the other two datasets correspond to news in Spanish, also properly labeled: \emph{The Spanish Fake News Corpus} \cite{posadas2019detection} containing 971 news items, and \emph{fake news in Spanish} \cite{bd_kaggle}, consisting of 1600 news items. None of the above datasets had missing or null data, and they are well balanced considering the two classes involved.
The English datasets were merged, resulting in a final English corpus comprising 51233 samples, of which 26645 are fake, and 24588 are genuine; the same procedure was followed over the individual datasets in Spanish, resulting in a final Spanish corpus consisting of 2571 samples, such that 1280 of them are fake and 1291 genuine \footnote{In this paper, we refer to the resulting datasets presented in this subsection as follows: the dataset in English, the dataset in Spanish, and the translated dataset.}. Figure \ref{fig:db_dist} shows through a chart the corresponding distribution for each resulting dataset.

Since the number of samples in Spanish is considerably small to train a DL model from scratch, one of the strategies followed during experiments, consists of evaluating the capacity of a DL model trained with the English corpus to predict fake Spanish news translated into English using the Google translation API.

\subsection{Experimental setup}
\label{sec:exp}

The experiments were carried out using a  \emph{Bootstrap (ShuffleSplit)} validation methodology, considering 5 iterations.

Depending on each scheme (described below), the partitioning was done into three subsets \emph{(train, development, test)} or into two subsets \emph{(train, test)}, taking 80\% for \emph{training} and 20\% for \emph{testing} (in the case of partitioning into three subsets, for the internal sub-division \emph{train/development} we took 80\% / 20\% respectively); however, in some experiments the split of the dataset was set to a ratio of \emph{train:} 90\% / \emph{test:} 10\%. These variations in the partitioning were considered due to the few samples we had in the final corpus in Spanish; hence, we wanted to try different combinations aiming at keep as many samples for training as possible.\\
       
Considering the balancing condition on the datasets we used, and the fact that we were addressing a bi-class classification task, we chose \emph{Accuracy} as the performance metric for measuring the generalization ability of the models.\\

Regarding the execution of experiments, four schemes were defined in combination with the datasets and the different models. Initially, a baseline was established, training and validating the four classical ML models listed in subsection \ref{sec:models}, using the dataset in Spanish \textbf{(first scheme)}. For classical ML approaches, the texts were represented using BoW and tf-idf techniques; this step was carried out to have a reference point for comparison purposes with the DL-based architectures. The subsequent experiments combine different DL models with different word embedding representations (presented in sub-section \ref{sec:models}), varying the datasets used for training and validation. The \textbf{second scheme} uses the dataset in Spanish both to train and validate two vanilla DL architectures based on LSTM and CNN layers, and more sophisticated architectures built on top of BERT-type models. Concerning the experiments with BETO embedding, we tried with different values for the number of epochs, and also applied the \emph{early stopping} strategy considering different values of the hyperparameters \emph{tolerance} and \emph{patience}. For its part, the \textbf{third scheme} is similar to the former one but using the dataset in English instead, so in this case no experiments were performed using the pre-trained BETO model. The last scheme trains the models with the dataset in English and validates with the translated dataset \textbf{(fourth scheme)}. Moreover, we conducted some experiments where samples from the translated dataset were progressively mixed with the dataset in English during the training phase; then, the remaining portion of the translated dataset was used for validation, i.e., emulating a learning curve.

The following is the collection of hyperparameter values we considered when training and validating the different models. Regarding the ML models for the baseline:

\begin{itemize}
  \item \textbf{(SVM)} RBF and Linear Kernel; regularization parameter 'C': 1e3, 1e-3; kernel coefficient for RBF 'gamma': 0.1, 1 
  \item \textbf{(RF and GBT)} number of trees: 50, 100, 200, 300, 500; maximum number of features: 50, 100, 200, 300
  \item \textbf{(MLP)} hidden layers: 1, 2, 3; number of neurons per hidden layer: 10, 50; epochs: 1000, 1500
\end{itemize} 

Furthermore, the combinations of these models were evaluated with BoW and tf-idf representations; removing and not removing Stop Words; applying and not applying Stemming; and considering a maximum vocabulary size of 10000, 20000, 30000 and 40000 words.

Similarly, for the DL models evaluated we considered:

\begin{itemize}
\item \textbf{LSTM:} units present in hidden layers (Units) [this model was only implemented with a single hidden layer], \emph{kernel regularizer} (KR), \emph{recurrent regularizer} (RR), \emph{dropout} (D).
\item \textbf{CNN:} amount of filters (F), \emph{kernel size} (KS), number of units for additional dense layer (Units), \emph{kernel regularizer} (KR).
\end{itemize}

It is also worth pointing out that, in order to set the input length for the models, we used a histograms-based approach to determine the most common length (in words) of the news items, in both the datasets for English and Spanish news: 1500 and 500 words, respectively.\\

The source code used to carry out the experiments can be found in a publicly accessible repository at GitHub \footnote{\url{https://github.com/kevinmaiden7/Spanish_FakeNewsDetection}}.

\subsection{Results}

Table \ref{tbl:baseline} shows the best results for each of the models considered during the experiments of the first scheme, which was described in subsection \ref{sec:exp}. It also shows the configuration of pre-processing steps that achieved the best results. Moreover, the hyperparameter values selected for each model were the following:

 \begin{itemize}
  \item \textbf{(SVM)} kernel RBF; 'C': 1e3; kernel coefficient 'gamma': 1 
  \item \textbf{(RF and GBT)} number of trees: 500; maximum number of features: 50
  \item \textbf{(MLP)} hidden layers: 1; neurons: 10; epochs: 1500
\end{itemize}

\begin{table}[!htbp]
    \centering
    \caption{Baseline results for the dataset in Spanish}
    \label{tbl:baseline}
    \begin{tabular}{|c|c|c|c|c|c|}
        \hline
        \textbf{Model} & 
        \textbf{Vocab Size} & 
        \textbf{Stemming} & 
        \textbf{Remove StopWords} & 
        \textbf{Text Representation} & 
        \textbf{test\_acc} \\   [1ex] 
        \hline
        SVM  & 10000 & NO & YES & tf-idf & 0.798 \\ 
        \hline
        RF & 40000 & NO & YES & tf-idf & 0.802 \\ 
        \hline
        GBT & 40000 & YES & NO & BoW & 0.783 \\ 
        \hline
        MLP & 10000 & YES & NO & tf-idf & 0.794  \\ 
        \hline
    \end{tabular}
\end{table}

According to the baseline results, RF in combination with a tf-idf text representation showed the highest \emph{accuracy}. Subsequently, we performed the experiments with the DL models (LSTM, CNN) in combination with the different types of Word Embeddings; hence, we followed the second, third, and fourth schemes. From this point on, we permanently removed Stop Words and did not apply Stemming anymore regarding data pre-processing.

Initially, we ran some experiments using a trainable embedding layer; the results are summarized in Table \ref{tbl:model_train_emb_sp_en}, where the hyperparameter values selected for each model were:

\begin{itemize}
    \item \textbf{LSTM (Spanish)} 16 units; KR and KK equals 1; D equals 0
    \item \textbf{LSTM (English)} 4 units; KR and KK equals 0.01; D equals 0
    \item \textbf{CNN (Spanish)} F equals 16; KS equals 10; 4 units; KR equals 0.01
    \item \textbf{CNN (English)} F equals 16; KS equals 10; 12 units; KR equals 0
\end{itemize} 

The LSTM and CNN models trained with the English dataset, whose results are shown in Table \ref{tbl:model_train_emb_sp_en}, were also validated with the whole translated dataset, yielding accuracies of 56.7\% and 53.2\% respectively.

\begin{table}[!htbp]
    \centering
    \caption{Results for DL models with trainable embedding layer; the column \textbf{dev\_acc} shows the \emph{accuracy} in the \emph{development} set; \textbf{std} is the standard deviation, and \textbf{test\_acc} shows the \emph{accuracy} in the \emph{test} set.}
    \label{tbl:model_train_emb_sp_en}
    \begin{tabular}{|c|c|c|c|c|}
        \hline
        \textbf{Model} & 
        \textbf{Dataset Language} &
        \textbf{dev\_acc} &
        \textbf{std} & 
        \textbf{test\_acc} \\ [1ex] 
        \hline
        LSTM & Spanish & 0.714 & 0.026 & 0.761 \\ 
        \hline
        LSTM & English & 0.95 & 0.02 & 0.931 \\
        \hline
        CNN & Spanish & 0.73 & 0.021 & 0.685 \\
        \hline
        CNN & English & 0.984 & 0.002 & 0.982 \\
        \hline
    \end{tabular}
\end{table}

Next, we performed the experiments using a Transfer Learning approach, with the pre-trained 300-feature GloVe embedding layer presented in subsection \ref{sec:models}; this time, the embedding values were left fixed during the training process, and only the added hidden layers were fine-tuned. Since the GloVe vectors utilized were trained on a corpus in English, these experiments correspond to the third and fourth schemes. The results are summarized in Table \ref{tbl:glove_eng}; moreover, the hyperparameter values chosen for each model were the following:

\begin{itemize}
  \item \textbf{(LSTM)} 8 units; KR and KK equals 0; D equals 0.5
   \item \textbf{(CNN)} F equals 16; KS equals 10; 4 units; KR equals 0
\end{itemize}

Then, when validating the former models using the translated dataset (fourth scheme), we got accuracy values of 54\% and 53.8\% for LSTM and CNN layers, respectively.

\begin{table}[!htbp]
    \centering
    \caption{Results for DL models using GloVe embedding with the dataset in English. The column \textbf{dev\_acc} shows the \emph{accuracy} in the \emph{development} set; \textbf{std} is the standard deviation, and \textbf{test\_acc} shows the \emph{accuracy} in the \emph{test} set.}
    \label{tbl:glove_eng}
    \begin{tabular}{|c|c|c|c|}
        \hline
        \textbf{Model} & 
        \textbf{dev\_acc} &
        \textbf{std} & 
        \textbf{test\_acc} \\ [1ex] 
        \hline
        LSTM  & 0.962 & 0.006 & 0.924 \\ 
        \hline
        CNN & 0.974 & 0.002 & 0.973 \\
        \hline
    \end{tabular}
\end{table}

\begin{figure}[!htbp]
    \centering
    \includegraphics[width=0.50\columnwidth]{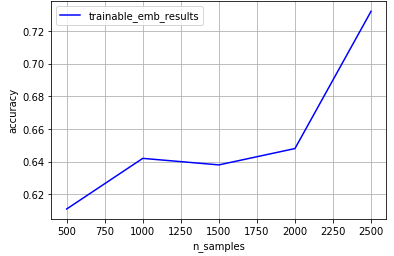}
    \caption{Learning curve results for CNN with trainable embedding}
    \label{fig:learning_curve}
\end{figure}

At this point, we decided to implement a learning curve in order to evaluate the effect of including data from the translated dataset, into a training dataset composed of the original news in English. The models evaluated included LSTM and CNN layers as well as trainable Embeddings and GloVe. Five experiments were run for each case, adding 500, 1000, 1500, 2000 and 2500 samples from the translated dataset to the training set, and then validating with the total amount of remaining samples of the translated dataset. Figure \ref{fig:learning_curve} shows the results for the CNN model with trainable Embedding, which was the best combination found in terms of accuracy, when applying this strategy.

Based on the previous results obtained with GloVe, and to simplify the experimental phase, the best hyperparameters of LSTM and CNN layers found using this embedding, were left fixed for the subsequent experiments applying ELMo and BERT-type embeddings.

\begin{table}[!htbp]
    \centering
    \caption{Results for ELMo, BERT and BETO Embeddings}
    \label{tbl:summary_emb}
    \begin{tabular}{|c|c|c|c|c|}
        \hline
        \textbf{Embedding} &
        \textbf{Model} & 
        \textbf{scheme} &
        \textbf{Epochs} &
        \textbf{test\_acc} \\ [1ex] 
        \hline
        ELMo  & LSTM & Third & 7 & 0.973 \\ 
        \hline
        ELMo  & CNN & Third & 5 & 0.957 \\ 
        \hline
        ELMo  & CNN & Fourth & 7 & 0.525 \\ 
        \hline
        BERT  & CNN & Third & 7 & 0.957 \\ 
        \hline
        BERT  & CNN & Fourth & 7 & 0.53 \\ 
        \hline
        BETO  & LSTM & Second & 25 & 0.80 \\ 
        \hline
    \end{tabular}
\end{table}

Regarding the results with ELMo embedding, we got high \emph{accuracy} outcomes for both LSTM and CNN models concerning the third scheme, as it is shown in Table \ref{tbl:summary_emb}; however, when it comes to the fourth scheme, the results show a degradation in performance. Additionally, taking into account the trend identified in the aforementioned learning curve approach, we added 2500 samples from the translated dataset to the training set, fitted the CNN layer in combination with ELMo embedding on this set, and validated with the remaining 71 samples of the translated dataset: we reached an \emph{accuracy} value of 70\%.\\

For the models built on top of BERT embedding, experiments were only carried out with the CNN layers; with this in mind, as it is described in Table \ref{tbl:summary_emb}, we obtained a high \emph{accuracy} result for the third scheme again, whereas not such a good level was achieved for the fourth scheme. Similarly to the procedure followed with ELMo embedding, 2500 samples were added from the translated dataset to the training set, for then validating with the remaining 71 samples: in this case we got an \emph{accuracy} level of 63.4\%.\\

\begin{figure}[!htbp]
    \centering
    \includegraphics[width=0.48\columnwidth]{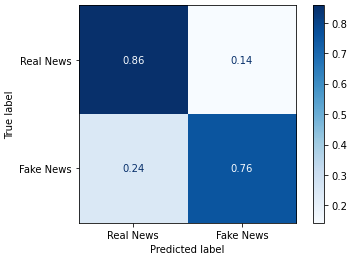}
    \caption{(BETO) normalized confusion matrix for LSTM with the dataset in Spanish}
    \label{fig:BETO_confusion_matrix}
\end{figure}

Concerning BETO embedding, the experiments with this model were framed in the second scheme (training and validating with the dataset in Spanish) since BETO itself is a model trained over a corpus in this language; in this case, we used both LSTM and CNN layers. Table \ref{tbl:summary_emb} shows the results for this experiment, where it is possible to observe that the LSTM model trained with 25 epochs achieved results of up to 80\% in \emph{accuracy} for the \emph{test} set; this is the best result we reached with DL models for the dataset in Spanish. Thus, Figure \ref{fig:BETO_confusion_matrix} shows the confusion matrix associated with this architecture for the \emph{test} set; it is worth highlighting that we used the early stopping strategy, although it did not yield a significant improvement in comparison to the previous results.

\section{Discussion and Conclusions}
\label{sec:discussion_and_conclusions}

Regarding the proposed baseline for datasets in Spanish (first scheme), RF showed the best performance getting an \emph{accuracy} of 80\%, and using the tf-idf text representation; however, this performance was not statistically different from that obtained with SVM, where a smaller vocabulary size was used. This model outperformed the best result reported in \cite{posadas2019detection}, where \emph{The Spanish Fake News Corpus} dataset was also utilized. Furthermore, we noticed that for this scheme, there were no significant differences in the performance of the models when applying Stemming or removing StopWords, or even when varying the text representation strategy or the vocabulary size.\\

It is worth highlighting the gap between the number of samples in the resulting datasets for English and Spanish, as it was shown in Figure \ref{fig:db_dist}. Since the models we used in this research follow a phenomenological approach, they highly depend on the amount of experimental data they are trained on. The above was evidenced by the prominent difference on \emph{accuracy} we obtained in the third and fourth schemes, using GloVe, ELMo, and BERT embeddings. Also, concerning the second scheme, we noticed that the models exhibited a trend of overfitting due to the small number of samples available for training; moreover, we observed that the regularization strategies we employed did not significantly improve the performance.\\

In regard to the fourth scheme, it is important to underline that the vocabulary present in the translated dataset corresponded to just 60\% of that present in the dataset in English; this situation negatively affected the results we obtained for the translation strategy. For this scheme, despite the excellent performance of the models trained and validated with the dataset in English (third scheme), when we validated with the translated dataset, the values of \emph{accuracy} were drastically reduced. Consequently, the results of the implemented learning curve indicated a performance improvement (although in different ratios) as more samples were added from the translated dataset to the training set (as it was shown in Figure \ref{fig:learning_curve}); this pattern was noticed regardless of the combination between a model and the embedding layer utilized.\\

Concerning the different embeddings we used, similar results were obtained for the third and fourth schemes when using GloVe, ELMo or BERT, in each of them; in fact, it is noteworthy that, in combination with these different embeddings, the LSTM and CNN layers showed similar results. Furthermore, taking into account these pre-trained models corresponded to the state-of-the-art in NLP, we were expecting to obtain salient results by mixing portions of the translated dataset to the dataset in English for training; however, due to the small number of samples available in the dataset in Spanish, and the discrepancies resulting from the translation, the predictive capability of the models was limited. Nonetheless, BETO embedding allowed us to obtain the best result for the dataset in Spanish (or any validation over news items in Spanish using DL models), considering the fact that its approximately 110 million parameters were pre-trained on a corpus in this language; this enabled us to take full advantage of the Transfer Learning strategy, and obtain an outstanding performance of 80\% of \emph{accuracy} on the \emph{test} set, in spite of the small number of samples available for such deep network architecture. Figure \ref{fig:BETO_confusion_matrix} showed the result for the LSTM model combined with this embedding, which corresponds to the best strategy identified for detecting fake news in Spanish using DL techniques: out of the 258 news items used for validation, the model correctly classified 76\% of the fake news items, as well as 86\% of the legitimate news items, which we consider a good hit ratio; by contrast, the model tends to confuse news items that are fake with legitimate ones, which corresponds to the main error condition it incurred.\\

Although the best detection rate achieved for DL models was similar to that obtained with RF, there is indubitably more room for improvement in the case of Deep Neural network architectures, due to the combination with Word Embeddings and more advanced techniques. Thus, in the future, this research could continue aiming at building a more robust system from the best strategy we found (BETO + LSTM), if a set of labeled news in Spanish with a more representative number of samples are available; furthermore, more experiments combining hyperparameter values and network architectures could also be carried out.

\bibliographystyle{ieeetr}
\bibliography{references}

\end{document}